\documentclass{article}
\usepackage{spconf,amsmath,graphicx,hyperref,caption,color,multirow}
\usepackage[export]{adjustbox}
\pdfoutput=1

\DeclareMathOperator*{\argmax}{arg\,max}

\newcommand{\klcomment}[1]{\textcolor{red}{\bf \small [ #1 --KL]}}
\newcommand{\llcomment}[1]{\textcolor{cyan}{\bf \small [ #1 --LL]}}
\newcommand{\kkcomment}[1]{\textcolor{blue}{\bf \small [ #1 --KK]}}
\newcommand{\kgcomment}[1]{\textcolor{orange}{\bf \small [ #1 --KG]}}
\renewcommand{\klcomment}[1]{\ignorespaces}
\renewcommand{\llcomment}[1]{\ignorespaces}
\renewcommand{\kkcomment}[1]{\ignorespaces}
\renewcommand{\kgcomment}[1]{\ignorespaces}

\newcommand{\kkfinal}[1]{\textcolor{blue}{\bf \small [ #1 --KK]}}

\renewcommand{\kkfinal}[1]{\ignorespaces}

\title{A study of all-convolutional encoders \\ for connectionist temporal classification}
\name{Kalpesh Krishna$^{\dagger}$\thanks{$^{\dagger}$ work done as a visiting student at TTIC}, Liang Lu$^{\dagger\dagger}$, Kevin Gimpel$^{\star}$, Karen Livescu$^{\star}$}
\address{$^\dagger$ Indian Institute of Technology Bombay, India, $^{\dagger\dagger}$ Microsoft AI and Research, USA \\$^{\star}$ Toyota Technological Institute at Chicago, USA\\\texttt{\{kalpesh,kgimpel,klivescu\}@ttic.edu, Liang.Lu@microsoft.com}}
\begin{document}
%
\maketitle
\begin{abstract}
Connectionist temporal classification (CTC) is a popular sequence prediction approach for automatic speech recognition that is typically used with models based on recurrent neural networks (RNNs).  We explore whether deep convolutional neural networks (CNNs) can be used effectively instead of RNNs as the ``encoder" in CTC.  CNNs lack an explicit representation of the entire sequence, but have the advantage that they are much faster to train.  We present an exploration of 
\klcomment{removed mention of 1-D and 2-D and added the following sentence} CNNs as encoders for CTC models, in the context of character-based (lexicon-free) automatic speech recognition.  In particular, we explore a range of one-dimensional convolutional layers, which are particularly efficient.  We compare the performance of our CNN-based models against typical RNN-based models in terms of training time, decoding time, model size and word error rate (WER) on the Switchboard Eval2000 corpus.  We find that our CNN-based models are close in performance to LSTMs, while not matching them, and are much faster to train and decode.\klcomment{added last sentence}
\end{abstract}
\begin{keywords}
Conversational speech recognition, connectionist temporal classification, convolutional neural networks, long short-term memory, lexicon-free recognition \llcomment{don't think we need this one}\kkcomment{isn't this an important aspect of the work?}
\end{keywords}
\section{Introduction}
\label{sec:intro}
In recent automatic speech recognition research, two 
types of neural models have become prominent:  recurrent neural network (RNN) encoder-decoders (``sequence-to-sequence" models) \cite{sutskever2014sequence,
chan2016listen,prabhavalkar2017comparison} and 
connectionist temporal classification (CTC) 
models \cite{graves2006connectionist,maas2015lexicon,amodei2016deep,miao2015eesen,zweig2017advances}. Both types of models perform well, but CTC-based models are more common in large state-of-the-art systems.  Among their advantages, CTC models are typically faster to train than encoder-decoders, because they lack the RNN-based decoder.

Most CTC-based models are based on variants of recurrent Long Short-Term Memory (LSTM) networks, sometimes including convolutional or fully connected layers in addition to the recurrent ones.
More recently, a few 
purely convolutional approaches to CTC \cite{zhang2017towards,wang2017residual} have been demonstrated to match or outperform LSTM counterparts. \klcomment{would be good to add a bit more detail.  in what way is our work different from this prior work on all-convolutional models?  can we say anything about why they outperform LSTMs while we don't?} \kkcomment{Both previous papers used 2-D convolutions. One of them just beat it on TIMIT, and I think we beat LSTMs on TIMIT as well. The other paper uses WSJ + their own dataset} \klcomment{interesting.  could we include TIMIT results?  it would be nice to show that we match prior work and then go beyond}\kkcomment{We got a 17.7\% dev PER on TIMIT with a CNN and 17.92\% using an LSTM. Zhang reports 17.4\% on a similar type of CNN, just 2D instead of 1D. I didn't run many LSTM experiments, but we could perhaps include just the CNN figure?} \klcomment{I could go either way.  The reason to include it is to show we can match prior work, so any deficit we find on Eval2000 is likely not because we are doing something wrong.  On the other hand some might see the 17.7\% and say we're not quite matching that 17.4\%} \kkcomment{The structure I'd used to get to 17.7 didn't have ResBlocks. Let's skip this, Zhang reports a 16.7\% as well using max-out non-linearity} \klcomment{OK} Purely convolutional networks have the advantage that they can be trained much faster, since all frames can be processed in parallel, whereas in recurrent networks the frames within an utterance cannot be naturally distributed across multiple processors.

We take a further step toward all-convolutional CTC models 
by exploring a variety of 
convolutional architectures trained with the CTC loss function and evaluating on conversational telephone speech (prior work evaluated on TIMIT, Wall Street Journal, and a corporate data set~\cite{zhang2017towards,wang2017residual}).  
Previous work with convolutional CTC models has mainly considered 2-D convolutional layers.  Here we study 1-D convolutions, which are more efficient and perform similarly. 1-D convolutions are similar to time-delay neural networks (TDNNs), which have traditionally been used with HMMs~\cite{waibel1990phoneme, peddinti2015time}.
\kkcomment{Ronan Collobert seems to have published 3-4 papers with the same CNN architecture, most recently~\cite{collobert2016wav2letter} (wav2letter) and ~\cite{palaz2015convolutional}} \klcomment{did some rephrasing to clarify we are only referring to CTC, and then added wav2letter mention in conclusion}

\kkfinal{Cut out a bit of text here to keep it to 4 pages, added TDNN just before this }
While the ideas should apply to any CTC-based model and task, here we consider the task of lexicon-free conversational speech recognition using character-based models.  We find that our best convolutional models are close to, but not quite matching, the best LSTM-based ones.  However, the CNNs can be trained much faster, so that given a fixed training time budget (within a wide range), convolutional models typically outperform recurrent ones. Our trained CNN models also convert speech-to-text much faster than their trained recurrent counterparts. As the research community considers increasingly large tasks, such as whole-word CTC models~\cite{audhkhasi2017direct,soltau2017neural}, computational efficiency is often a concern, especially with limited hardware resources. The efficiency of CNNs makes them an attractive option in these settings.\klcomment{added last sentence}\kkcomment{}
\begin{figure}
\caption{CNN encoders, with filter size noted in each block. We have tuned $N$ and $K$ (time filter size) in our experiments.\klcomment{removed tensorflow mention to save a line :)}\kkcomment{Changed it to 1-D only figure, change filename to network for prev figure}}
\label{fig:network1d}
\includegraphics[scale=0.6, center]{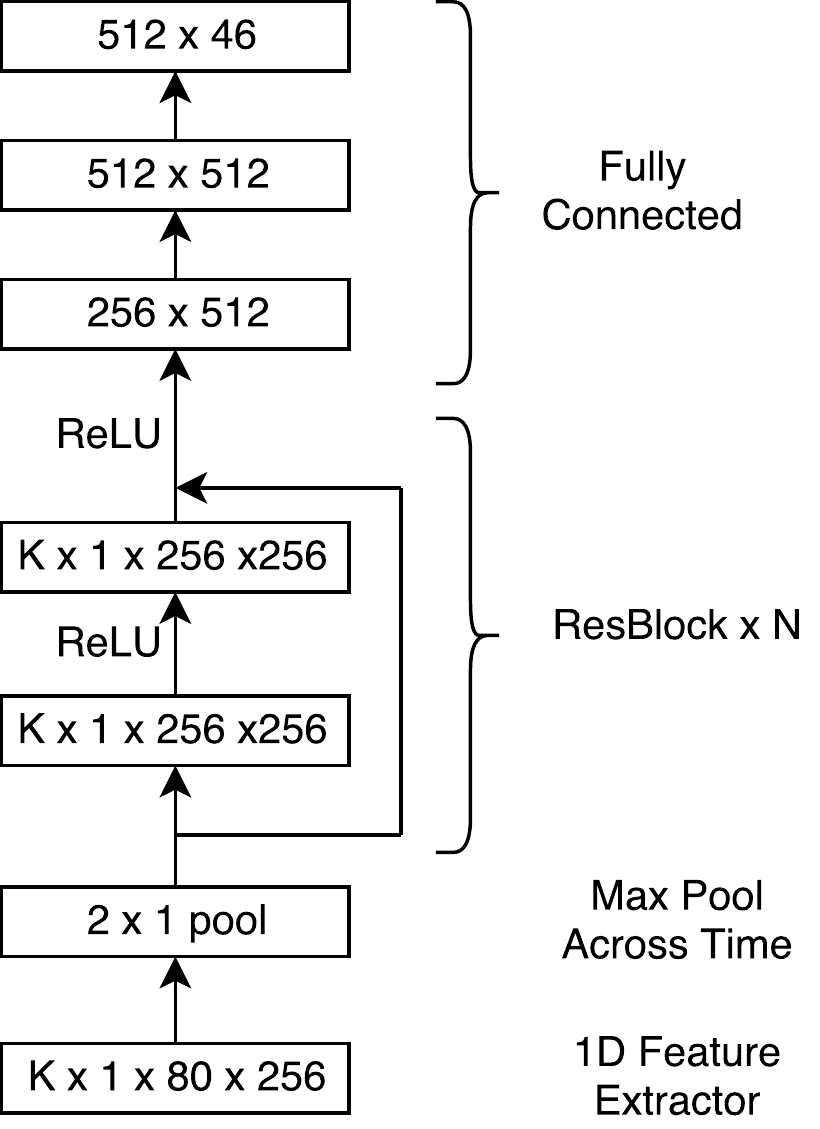}
\end{figure}

\section{Model Architecture}
\label{sec:pagestyle}
\vspace{-.05in}
CTC is an approach to sequence labeling that uses a neural ``encoder'', which maps from an input sequence of frame features $\textbf{x} = (x_1, x_2, \ldots, x_T)$ to a sequence of hidden state vectors $h_t$, followed by a softmax to produce posterior probabilities of frame-level labels (referred to as ``CTC labels") $p(\pi_t|h_t)$ for each label $\pi_t \in \mathcal{C}$.  The posterior probability of a complete frame-level label sequence is taken to be the product of the frame posteriors:
\vspace{-.05in}
\begin{equation}
p(\pi = \pi_1, \pi_2, \ldots, \pi_T | \mathbf{x}) = \prod_t p(\pi_t|h_t)
\end{equation}
The CTC label set $\mathcal{C}$ consists of all of the possible true output labels (in our case, characters) plus a ``blank" symbol $\phi$.  Given a CTC label sequence, the hypothesized final label sequence is given by collapsing consecutive identical frame CTC labels and removing blanks.  We use $B(\pi)$ to denote the collapsing function. \klcomment{edited slightly}
All of the model parameters are learned jointly using the CTC loss function, which is the log posterior probability of the training label sequence $\textbf{z} = z_1, z_2 \ldots, z_L$ given input sequence $\mathbf{x}$,
\begin{eqnarray}
\log p(\mathbf{z} | \mathbf{x}) &=& \log \sum_{\pi \in B^{-1}(\mathbf{z})} p(\pi | \mathbf{x})\\
&=& \log \sum_{\pi \in B^{-1}(\mathbf{z})} \prod_t p(\pi_t | h_t)
\end{eqnarray}
Model parameters are learned using gradient descent; the gradient can be computed via a forward-backward technique~\cite{graves2006connectionist}.

\vspace{-.1in}
\subsection{Decoding}
\label{sec:decoding}

Our CTC 
models operate at a character level.  We use the special blank symbol $\phi$ along with a vocabulary of 45 characters which appear in the raw SWB corpus (26 letters, 10 digits, space, \&, ', -, [\textit{laughter}], [\textit{vocalized-noise}], [\textit{noise}], / and \_). These transcriptions were inherited from a Switchboard Kaldi~\cite{povey2011kaldi} setup without text normalization. We remove punctuation and noise tokens during post-processing. \klcomment{say what the 45 characters are. also, by '\_' do you mean space?} \kkcomment{No, I referred to the special symbol used by CTC} \klcomment{I think we should use one of the commonly used names for blank, either $\phi$ or {\it blank}}\kkcomment{} \llcomment{could mention that the transcriptions were inherited from the Kaldi baseline without further text normalization.}\kkcomment{added this} Decoding with CTC models can be done in a number of ways, depending on whether one uses a lexicon and/or a word- or character-level language model (LM)~\cite{zenkel2017comparison}.  Here we focus on two simple cases, greedy decoding with no language model and beam-search decoding with an $n$-gram character LM.

To decode without a language model, we take the most likely CTC output label at each frame and collapse the resulting frame label sequence to the corresponding character sequence.
We also consider decoding with an $n$-gram language model ($n=7,9$) using a beam search decoding procedure. We decode with the objective,
 \begin{align*}
\mathbf{\hat{\pi}} = \argmax_{\mathbf{\pi} \in \mathbf{\Pi}}~~
p(\mathbf{k})^{\alpha}|\mathbf{k}|^{\beta} \prod_t p(\pi_t|h_t)
\end{align*}
where $\mathbf{k} = B(\mathbf{\pi})$,  $|\mathbf{k}|$ denotes the length of $\mathbf{k}$, and $\alpha$, $\beta$ are tunable parameters. The final decoded output is $\mathbf{\hat{z}} = B(\mathbf{\hat{\pi}})$. 
Our beam-search method is the algorithm described in~\cite{maas2015lexicon}.\footnote{We account for \texttt{<s>} and \texttt{</s>} tokens during beam-search decoding (not explicitly mentioned in the beam search algorithm in  \cite{maas2015lexicon}).}\klcomment{do you mean that maas et al.~didn't properly account for these tokens?  If so, what are we doing differently?}\kkcomment{Algorithm 1 in Maas et al. starts each beam with an empty string, whereas we are starting with a \texttt{<s>} token and not including it in $|k|$. Once our beam search ends, we re-score the beam by adding LM probability of \texttt{</s>} at the end of of each beam element. While Maas et al. does mention the use of \texttt{<s>} / \texttt{</s>} in their language model, they do not specify this in Algorithm 1. These two changes were improving results by \~2\%} \llcomment{This detail is intresting, and could be presented in the paper.}\klcomment{I edited slightly. no opinion on expanding on this point}\kkcomment{Added as a footnote}
\subsection{Encoders}
We refer to the neural network that maps from the input 
$\textbf{x}$ to 
state vectors $\textbf{h}$ as an encoder.  We consider both a typical recurrent LSTM encoder and various convolutional encoders.  Our input 
vectors are 40 log mel frequency filterbank features (static) concatenated with their first-order derivatives (delta).
\vspace{-.05in}
\subsubsection{LSTMs}
Our 
recurrent encoder is a multi-layer bi-directional LSTM with a dropout layer 
between consecutive layers (with dropout rate 0.1). We concatenate every two consecutive input vectors 
(as in \cite{chan2016listen}),
which reduces the time resolution by a factor of two and speeds up both the forward and backward pass.
\vspace{-.05in}
\subsubsection{1-D CNNs}
For our all-CNN encoders, we consider 1-D CNN structures that convolve across time only.  Each of the input acoustic feature dimensions is treated as a separate input channel.  
The first layer is a convolution followed by max-pooling across time (with a stride size 2), followed by several convolutional layers, and ending with two 512-unit fully connected layers and a final projection layer. Each convolution has 256 channels.
We add batch normalization after every convolution, and include residual connections between every pair of convolutional layers after the max-pool \cite{he2016deep, ioffe2015batch}. A ReLU  \cite{nair2010rectified} non-linearity is used after every convolution, similar to the residual learning blocks in \cite{he2016deep} (referred to as ``ResBlocks (RBs)'' in the rest of the paper). \textbf{\hyperref[fig:network1d]{Fig. \ref{fig:network1d}}} portrays our architecture. 

\section{Experimental Setting}
\label{sec:typestyle}
\vspace{-.05in}
\subsection{Data Setup}
\vspace{-.05in}
We use the Switchboard corpus (LDC97S62) \cite{godfrey1992switchboard}, which contains roughly 300h of conversational telephone speech, as our training set. Following the Kaldi recipe~\cite{povey2011kaldi}, we reserve the first 4K utterances as a validation set. Since the training set has several repetitions of short utterances (like ``uh-huh''), we remove duplicates beyond a count threshold of 300. The final training set has about 192K utterances. For evaluation, we use the HUB5 Eval2000 data set (LDC2002S09), consisting of two subsets: Switchboard (SWB), which is similar in style to the training set, and CallHome (CH), which contains conversations between friends and family.\footnote{Our Eval2000 setup has 4447 utterances, 11 utterances fewer than in some other papers. This discrepancy could result in an Eval2000 WER difference of 0.1-0.2\%.} \klcomment{edited footnote a bit}
Our input filterbank features along with their deltas are normalized with per-speaker mean and variance normalization.
\vspace{-.05in}
\subsection{Training Setup}
All models are trained on a single Titan X GPU with two supporting CPU threads, using TensorFlow \texttt{r1.1} \cite{abadi2016tensorflow} and optimized using Adam \cite{kingma2014adam} with a mini-batch size of 64 for LSTM (\texttt{BasicLSTMCell})  models and 32 for CNN models (unless otherwise mentioned). For the LSTM models, we use a learning rate of 0.001. For the CNN models, a smaller learning rate of 0.0002 was preferred. The learning rate is decayed by 5\% whenever validation loss doesn't decrease over two epochs. We report average training time per epoch for each model as both wall-clock hours ($t_{wc}$) and CPU-hours ($t_{cpu}$).


\section{Results}
\vspace{-.05in}
\subsection{LSTM Baseline}
\vspace{-.05in}
\kkcomment{Decided to remove the LSTM baseline experiments. Might as well just stick to the model in Zenkel et al. Some other LSTMs were coming daringly close, and it's best we don't try to counteract existing literature on LSTMs since it's not the focus of the paper}
As a baseline, we train a 5-layer 320 hidden unit bi-directional recurrent neural network using LSTMs, similar to the architecture described in~\cite{zenkel2017comparison}. With a batch-size of 64, our LSTM needs $t_{wc} = 3.3$ hours / epoch and $t_{cpu} = 5.8$ hours / epoch. On a batch-size of 32, the LSTM takes $t_{wc} = 8.7$ hours / epochs and $t_{cpu} = 14.8$ hours / epoch.
\begin{table}
\begin{center}
\caption{
Development set WER for 1-D CNNs vs.~number of layers. $b$ denotes batch-size. Each model is trained for 40 epochs with early stopping. $t_{wc} / t_{cpu}$ are hours / epoch.}
\label{tab:conv}
\begin{tabular}{ |l|r|c|c|c| } 
 \hline
Model & \# Weights & WER \% & $b$ & $t_{wc} / t_{cpu} (h)$\\
\hline
5/320 LSTM & 11.1M & 28.54 & 64 & 3.3 / 5.8 \\
\hline
10*1, 8 RBs & 11.1M & 36.71  & 32 & 0.9 / 2.2\\
10*1, 11 RBs & 15.1M & 32.67 & 32 & 1.0 / 2.5\\
10*1, 14 RBs & 19.0M & 30.92 & 32 & 1.1 / 2.8\\
10*1, 17 RBs & 22.9M & \textbf{29.82} & 32 & 1.5 / 3.5\\
\hline
\end{tabular}
\end{center}
\end{table}

\vspace{-.05in}
\subsection{1-D CNNs}
\begin{figure}
\caption{Comparison of convergence vs.~wall-clock time. 5/320 denotes the 5-layer 320-unit LSTM, $b$ = batch size.\klcomment{edited for length}
}
\label{fig:time}
\vspace{-.05in}
\includegraphics[scale=0.6, center]{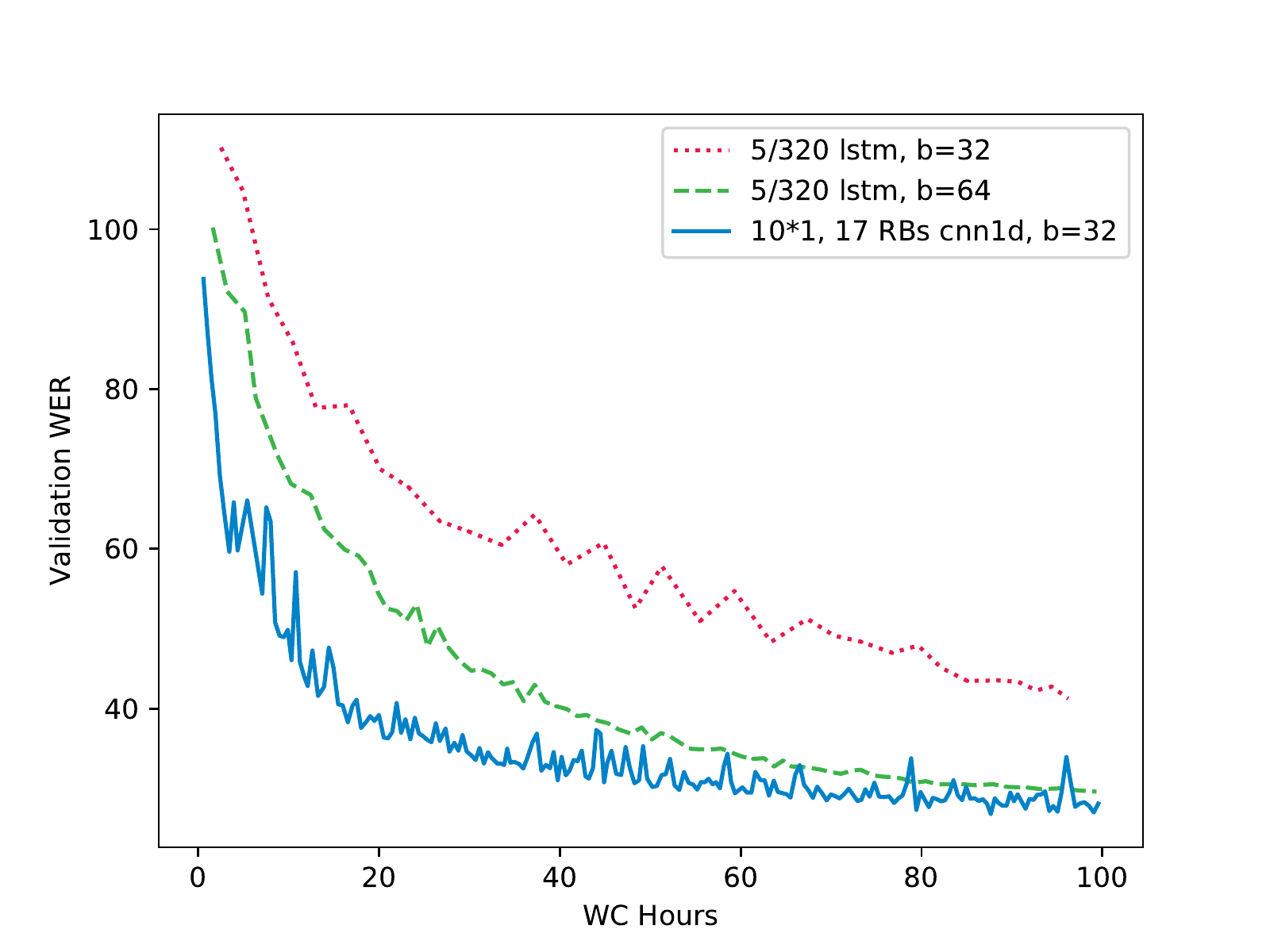}
\end{figure}

\vspace{-.05in}
We conduct experiments on 1-D CNNs investigating variance in performance and time / epoch with network depth and filter size. These are given in \textbf{\hyperref[tab:conv]{Table \ref{tab:conv}}} and \textbf{\hyperref[tab:filter]{Table \ref{tab:filter}}}. We notice that for the same number of trainable parameters deeper networks with smaller filters seem to perform the best. We noticed that smaller-filter deeper architectures over-fit less when compared to larger-filter architectures with the same number of trainable parameters. For a fixed network depth, a mid-sized filter performed best. We present a graph of convergence vs wall-clock time in \textbf{\hyperref[fig:time]{Fig. \ref{fig:time}}}. As expected, the CNNs train faster than LSTMs, and significantly faster at the same batch-size. We also notice significant speed-ups during greedy decoding of the Eval2000 corpus, as shown in \textbf{\hyperref[tab:decode]{Table \ref{tab:decode}}}.

\begin{figure}
\caption{Visualization of the filters in the first layer for the 10*1, 17 RBs 1-D CNN. For each filter, the horizontal axis represents time and the vertical axis represents the 80 input channels (40 static + 40 delta, separated by the checkerboard pattern). The $6^{th}$ and $7^{th}$ images are a $\max$ and $\min$ over all 256 output channels.}
\label{fig:filter}
\begin{center}
\includegraphics[scale=0.28]{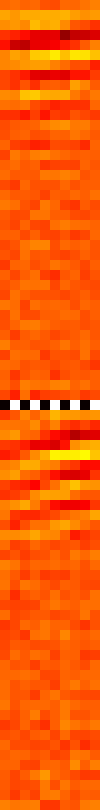}
\includegraphics[scale=0.28]{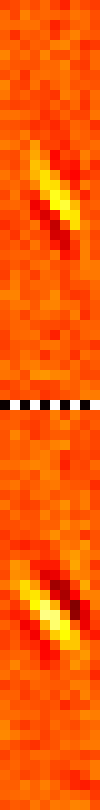}
\includegraphics[scale=0.28]{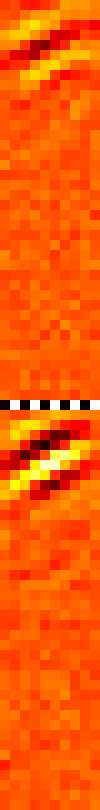}
\includegraphics[scale=0.28]{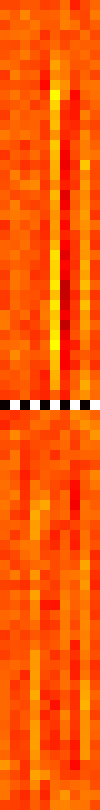}
\includegraphics[scale=0.28]{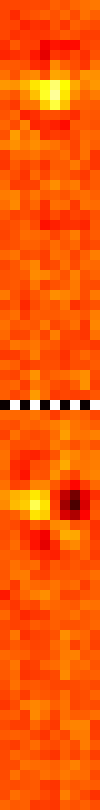}
\includegraphics[scale=0.28]{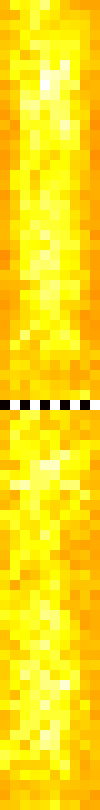}
\includegraphics[scale=0.28]{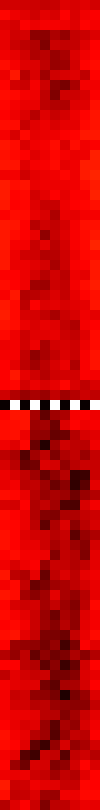}
\end{center}
\end{figure}

We show some of the learned filters in \textbf{\hyperref[fig:filter]{Fig.  \ref{fig:filter}}}. These filters show that the network learns derivative-like filter patterns  \kkcomment{I meant ``difference filters, [-1, 0, 1]'' here} across different input channels. 
Our 1-D convolution structure with filter size $K$*1 can be viewed as similar to a 2-D convolution with filter size $K$*80, since the 1-D filters are learned jointly.  \klcomment{reworded a bit} We also note the strong relation between filter patterns learned in the static and delta regions.


\begin{table}
\begin{center}
\caption{Development set WER for 1-D CNNs vs.~filter size, each trained for 40 epochs with early stopping. The first two experiments vary filter size / depth at a constant number of trainable parameters (approximately for 15*1 filter). The third experiment varies filter size at a constant depth.
}
\label{tab:filter}
\begin{tabular}{ |l|r|c|c|c|c| } 
 \hline
Model & \# Weights & WER \% & $t_{wc} / t_{cpu} (h)$\\
\hline
5*1, 16 RBs & 11.1M & \textbf{33.26} & 1.0 / 2.3 \\
10*1, 8 RBs & 11.1M & 36.71 & 0.9 / 2.2 \\
15*1, 5 RBs & 10.5M & 43.18 & 0.8 / 2.1 \\
15*1, 6 RBs & 12.4M & 39.83 & 0.9 / 2.4 \\
\hline
5*1, 28 RBs & 19.0M & \textbf{29.65} & 1.4 / 3.5  \\
10*1, 14 RBs & 19.0M & 30.92 & 1.1 / 2.8\\
15*1, 9 RBs & 18.3M & 35.45 & 1.1 / 3.1 \\
15*1, 10 RBs & 20.3M & 33.94 & 1.1 / 3.0 \\
\hline
5*1, 14 RBs & 9.8M & 35.34 & 1.0 / 2.2 \\
10*1, 14 RBs & 19.0M & \textbf{30.92} & 1.1 / 2.8 \\
15*1, 14 RBs & 28.1M & 31.36 & 1.6 / 3.8 \\
\hline
\end{tabular}
\end{center}
\end{table}

\begin{table}
\begin{center}
\caption{Greedy decoding time on the Eval2000 corpus (4447 utterances). $b$ (batch-size) $=1$ is practical in real-time systems since it decodes one utterance at a time. $t_{wc} / t_{cpu}$ represent total decoding time in \textbf{seconds} averaged over three runs.}
\label{tab:decode}
\begin{tabular}{ |l|r|r|rcrl| } 
 \hline
Model & \# Weights & $b$ & $t_{wc}$ \hspace{-.15in} &/& \hspace{-.15in} $t_{cpu}$ \hspace{-.15in}& \hspace{-.15in}$(s)$\\
\hline
5/320 LSTM & 11.1M & 1 & 1813 \hspace{-.15in} &/& \hspace{-.1in}3667 &\\
5/320 LSTM & 11.1M & 32 & 87 \hspace{-.15in} &/& \hspace{-.1in}180 &\\
5/320 LSTM & 11.1M & 64 & 44 \hspace{-.15in} &/& \hspace{-.1in}92 &\\
\hline
5*1, 28 RBs, CNN & 19.0M & 1 & 115 \hspace{-.15in} &/& \hspace{-.1in}135 &  \\
5*1, 28 RBs, CNN & 19.0M & 32 & 17 \hspace{-.15in} &/& \hspace{-.1in}18 &\\
5*1, 28 RBs, CNN & 19.0M & 64 & 15 \hspace{-.15in} &/& \hspace{-.1in}16 &\\
\hline
\end{tabular}
\end{center}
\end{table}
\vspace{-.05in}
\subsection{Language Model Decoding}
\label{sec:lmdecoding}
\vspace{-.05in}
We evaluate our baseline LSTM and best performing CNN (5*1 filter with 28 RBs) on the Eval2000 corpus. We train each model to 50 epochs with early stopping on validation data.
We augment our models with 7-gram and 9-gram character-level language models (LMs). These $n$-gram models were trained only on the SWB \klcomment{choose either SWB or SWBD and use it throughout (currently there is a mix)}\kkcomment{} training corpus transcripts using SRILM \cite{stolcke2002srilm}. For all experiments, a beam size of 200 was used. We choose $\alpha = 0.6$ and $\beta = 1.5$ after tuning on validation data. Our results are presented in \textbf{\hyperref[table:eval2000]{Table \ref{table:eval2000}}}. Notice that in the no LM results our CNNs are only 0.2\% behind on the SWB part of Eval2000, but a larger 1.1\% behind on CH. After LM decoding, the differences are more pronounced. This indicates that CNNs seem to over-fit more on the training data (which is similar to the SWB part of Eval2000) and show less improvement with the help of LMs.  

\begin{table}
\caption{
Final test set results on Eval2000.
\klcomment{this table has a lot of results for a "final results" table.  I suggest paring it down, perhaps keeping only the final best results plus any others that are needed for comparison with prior work. also, list all prior work in a single table section rather than two?}\kkcomment{Moved the other baselines down to make the comparison with Maas et al. clearer?}\klcomment{OK.  remove 5-g results?}\kkcomment{}}
\label{table:eval2000}
\vspace{-.1in}
\begin{center}
\begin{tabular}{ |l|r|r|r| } 
 \hline
 Model& SWB & CH & EV \\ 
 \hline
5/320 LSTM + no LM& \textbf{27.7} & \textbf{47.5} & 37.6 \\
5/320 LSTM + 7-g & 20.0 & 38.5 & 29.3 \\
5/320 LSTM + 9-g & \textbf{19.7} & \textbf{38.2} & 29.0 \\
\hline
5*1 28 RBs, CNN + no LM & \textbf{27.9} & \textbf{48.6} & 38.3 \\
5*1 28 RBs, CNN + 7-g & 21.7 & 40.4 & 31.1 \\
5*1 28 RBs, CNN + 9-g & \textbf{21.3} & \textbf{40.0} & 30.7 \\
\hline
Maas \cite{maas2015lexicon} + no LM & 38.0 & 56.1 & 47.1 \\
 Maas \cite{maas2015lexicon} + 7-g & 27.8 & 43.8 & 35.9 \\
 Maas \cite{maas2015lexicon} + RNN & 21.4  & 40.2 & 30.8 \\
 \hline
 Zenkel \cite{zenkel2017comparison} + no LM & 30.4 & 44.0 & 37.2 \\
 Zenkel \cite{zenkel2017comparison} + RNN & 18.6 & 31.6 & 25.1\\
 \hline
 Zweig \cite{zweig2017advances} + no LM & 25.9 & 38.8 & -\\
 Zweig \cite{zweig2017advances} + $n$-g & 19.8 & 32.1 & -\\
\hline 
\end{tabular}
\end{center}
\end{table}

\vspace{-.05in}
\section{Conclusions}
\vspace{-.05in}
We take a further step towards making all-convolutional CTC architectures practical for speech recognition.
In particular we have explored 1-D convolutions with CTC, which are particularly time-efficient. 
\kkcomment{Maybe not very memory efficient}\klcomment{good point.  I meant relative to 2-D CNN, but that may not be a selling point compared to LSTMs...}
Our CNN-based CTC models are still slightly behind LSTMs in performance, but train and decode significantly faster. Further work in this space could include additional model variants and regularizers, as well as studying the relative merits of all-convolutional models in larger systems operating at the word level, where the efficiency advantages are expected to be even more important. In addition, CNN-based speech recognition has also been explored in the context of different training and decoding algorithms, such as the auto segmentation criterion~\cite{collobert2016wav2letter}. It would be interesting to conduct a broader study considering the interaction of CNNs with different training and decoding approaches.
\klcomment{references could use some cleanup, and could be shortened by removing page numbers, using abbreviations, etc.}\kkcomment{Removed page numbers, what can we abbreviate?}\klcomment{no need to shorten once it fits on 1 page, but a few cleanup items:  make references to the same conference consistent (e.g. including "Proceedings" or not, spelling out vs. abbreviating, etc.), update arxiv preprints to their conference/journal form when it exists (e.g. Listen attend and spell paper), clean up refs that mention the year or IEEE multiple times, check capitalization}
\vspace{-.05in}
\section{Acknowledgements}
\vspace{-.05in}
We are grateful to Shubham Toshniwal for help with the data and baselines, and to Florian Metze for useful comments.

\vfill\pagebreak
\bibliographystyle{IEEEbib}
\bibliography{refs}{}

\end{document}